\documentclass{article}

\usepackage{microtype}
\usepackage{graphicx}
\usepackage{subfigure}
\usepackage{booktabs} % for professional tables
\usepackage{mathpartir}
\usepackage{makecell}

\usepackage{hyperref}
\usepackage{breakurl}

\usepackage[accepted]{icml2019}

\icmltitlerunning{Neural Rewriting}

\begin{document}

\twocolumn[
	\icmltitle{
	Can Neural Networks Learn Symbolic Rewriting?
	}



\icmlsetsymbol{equal}{*}

\begin{icmlauthorlist}
\icmlauthor{Bartosz Piotrowski}{mimuw,ciirc}
\icmlauthor{Josef Urban}{ciirc}
\icmlauthor{Chad E. Brown}{ciirc}
\icmlauthor{Cezary Kaliszyk}{innsbruck,mimuw}
\end{icmlauthorlist}

\icmlaffiliation{ciirc}{Czech Technical University, Prague}
\icmlaffiliation{mimuw}{University of Warsaw, Poland}
\icmlaffiliation{innsbruck}{University of Innsbruck, Austria}

\icmlcorrespondingauthor{Bartosz Piotrowski}{bartoszpiotrowski@post.pl}
\icmlcorrespondingauthor{Josef Urban}{josef.urban@gmail.com}

\icmlkeywords{Machine Learning, Deep Learning, Neural Machine Translation,
Symbolic Rewriting, ICML}

\vskip 0.3in
]



\printAffiliationsAndNotice{\icmlEqualContribution} % otherwise use the standard text.

\begin{abstract}
	This work investigates if the current neural architectures are adequate for
	learning symbolic rewriting. Two kinds of data sets are proposed for this
	research -- one based on automated proofs and the other being a synthetic
	set of polynomial terms. The experiments with use of the current neural
	machine translation models are performed and its results are discussed.
	Ideas for extending this line of research are proposed, and its relevance
	is motivated.
\end{abstract}

\section{Introduction}
\label{sect:intro}

Neural networks (NNs) turned out to be very useful in several domains. In
particular, one of the most spectacular advances achieved with use of NNs has
been natural language processing. One of the tasks in this domain is a
translation between natural languages -- neural machine translation (NMT)
systems established here the state-of-the-art performance. Recently, NMT
produced first encouraging results in the \textit{autoformalization
task}~\cite{KaliszykUVG14,KaliszykUV15,KaliszykUV17,Wang18} where given an
\textit{informal} mathematical text in \LaTeX{} the goal is to translate it to
its \textit{formal} (computer understandable) counterpart. In particular, the
NMT performance on a large synthetic \LaTeX{}-to-Mizar dataset produced by a
relatively sophisticated toolchain developed for several
decades~\cite{BancerekNU18} is surprisingly good~\cite{Wang18}, indicating that
neural networks can learn quite complicated algorithms for symbolic data. This
inspired us to pose a question: \textit{Can NMT models be used in the
formal-to-formal setting?} In particular: \textit{Can NMT models learn symbolic
rewriting?}

The answer is relevant to various tasks in automated reasoning. For example,
neural models could compete with symbolic methods such as inductive logic
programming~\cite{muggleton1994inductive} (ILP) that have been previously
experimented with to learn simple rewrite tasks and theorem-proving heuristics
from large formal corpora~\cite{urban98}. Unlike (early) ILP, neural methods
can, however, easily cope with large and rich datasets without combinatorial
explosion.

Our work is also an inquiry into the capabilities of NNs as such, in
the spirit of works like \cite{Evans18}.

% for this is not of a completely direct relevance for automated
% reasoning -- this is rather an inquiry on the capabilities of NNs as such, in a
% spirit of works like \cite{Evans18}. Nevertheless, there is a high chance that
% various symbolic reasoning tasks, difficult to solve by classical methods, can
% be cast as machine learning problems, also NMT-like ones \cite{Sekiyama17}.
% The insights on how NMT models work in a completely symbolic setting gained here
% can be important for the problems important for enhancing automated reasoning.

\section{Data}
\label{sect:data}

To perform experiments answering our question, we prepared two data sets -- the
first consists of examples extracted from proofs found by ATP (automated
theorem prover) in a mathematical domain (AIM loops), whereas the second is a
synthetic set of polynomial terms.

\subsection{The AIM data set}

The data consists of sets of ground and nonground rewrites that came from % real
\texttt{Prover9}\footnote{\url{https://www.cs.unm.edu/~mccune/prover9/}}
proofs of theorems about AIM loops produced by Veroff~\cite{Veroff13}.

Many of the inferences in the proofs are paramodulations from an equation and
have the form
\begin{mathpar}
	\inferrule{s = t \\ u[\theta(s)] = v}{u[\theta(t)] = v}
\end{mathpar}
where $s, t, u, v$ are terms and $\theta$ is a substitution. For the most
common equations $s = t$, we gathered corresponding pairs of terms
$\big(u[\theta(s)], u[\theta(t)]\big)$ which were rewritten from one to another
with $s = t$. We put the pairs to separate data sets (depending on the
corresponding $s = t$): in total 8 data sets for
ground rewrites (where $\theta$ is trivial) and 12 for nonground ones.
The goal will be to learn rewriting for each of these 20 rules separately.

Terms in the examples are treated as linear sequences of tokens where tokens
are single symbols (variable / constant / predicate names, brackets, commas).
Numbers of examples in each of the data sets vary between 251 and 34101.
Lengths of the sequences of tokens vary between 1 and 343, with the mean around
35.  These 20 data sets were split into training, validation and test sets for
our experiments ($60 \%, 10 \%, 30 \%$, respectively).

In Table \ref{tab:aim_ground} and Table \ref{tab:aim_nonground} there are
presented examples of pairs of AIM terms in TPTP \cite{Sutcliffe17} format,
before and after rewriting with, respectively, ground and nonground rewrite
rules.\footnote{All the described AIM data are available at
\url{https://github.com/BartoszPiotrowski/rewriting-with-NNs/tree/master/data/AIM}}

%\begin{table}[]
%\caption{Examples in the AIM data set.}
%\label{tab:aim}
%\begin{tabular}{l||l|l}
%	Rewrite rule: & Before rewriting: & After rewriting: \\
%	\hline
%	\texttt{b(s(e,v1),e) = v1} &
%	\texttt{k(b(s(e,v1),e),v0)} & % Chad: Changed some of these from v0 to v1, since this must be ground rewrite
%	\texttt{k(v1,v0)} \\ % Chad: Changed some of these from v0 to v1, since this must be ground rewrite
%	\texttt{o(V0,e) = V0} &
%	\texttt{t(v0,o(v1,o(v2,e)))} &
%	\texttt{t(v0,o(v1,v2))} \\
%	\texttt{k(V0,k(V1,V2)) = k(V1,k(V0,V2))} &
%	\texttt{l(k(v1,k(v0,v2)),k(v0,v2),v3)} &
%	\texttt{l(k(v0,k(v1,v2)),k(v0,v2),v3)}
%\end{tabular}
%\end{table}

\begin{table}[]
\caption{Example of a ground rewrite in the AIM data set.}
\label{tab:aim_ground}
\begin{tabular}{l|l}
	Rewrite rule: & \texttt{b(s(e, v1), e) = v1} \\
	Before rewriting: & \texttt{k(b(s(e, v1), e), v0)} \\
	After rewriting: & \texttt{k(v1, v0)}
\end{tabular}
\end{table}

\begin{table}[]
\caption{Example of a nonground rewrite in the AIM data set.}
\label{tab:aim_nonground}
\begin{tabular}{l|l}
	Rewrite rule: & \texttt{o(V0, e) = V0} \\
	Before rewriting: & \texttt{t(v0, o(v1, o(v2, e)))} \\
	After rewriting: & \texttt{t(v0, o(v1, v2))}
\end{tabular}
\end{table}

\subsection{The polynomial data set}

This is a synthetically created data set where the examples are pairs of
equivalent polynomial terms. The first element of each pair is a polynomial in
an arbitrary form, and the second element is the same polynomial in a normalized
form. The arbitrary polynomials are created randomly in a recursive manner
from a set of available (non-nullary) function symbols, variables and
constants. First, one of the symbols is randomly chosen. If it is a constant or
a variable, it is returned and the process terminates. If a function symbol is
chosen, its subterm(s) are constructed recursively in a similar way.

\begin{table}[]
\caption{Examples in the polynomial data set.}
\label{tab:poly}
\begin{tabular}{l|l}
	Before rewriting: & After rewriting: \\
	 \hline
	\texttt{(x*(x+1))+1}&
	\texttt{x\^{}2+x+1}\\
	\texttt{(2*y)+(1+(y*y))}&
	\texttt{y\^{}2+2*y+1}\\
	\texttt{(x+2)*(((2*x)+1)+(y+1))}&
	\texttt{2*x\^{}2+5*x+y+3}
\end{tabular}
\end{table}

The parameters of this process are set in such a way that it creates polynomial
terms of average length around 25 symbols. Terms longer than 50 are filtered
out.  Several data sets of various difficulties were created by varying the
number of available symbols. These were quite limited -- at most 5 different
variables and constants being a few first natural numbers.  The reason for this
limited complexity of the input terms is because normalizing even a relatively
simple polynomial can result in a very long term with very large constants --
which is related especially to the operation of exponentiation in polynomials.

Each data set consists of different 300 000 examples -- see Table \ref{tab:poly}
for examples. These data sets were split into training, validation and test
sets for our experiments ($60 \%, 10 \%, 30 \%$, respectively).\footnote{The
described polynomial data are available at
\url{https://github.com/BartoszPiotrowski/rewriting-with-NNs/tree/master/data/polynomial}}
% TODO check this link

\section{Experiments}
\label{sect:exp}

For experiments with both data sets, we used an established NMT architecture
\cite{Luong17} based on LSTMs (long short-term memory cells) and implementing
the attention mechanism.\footnote{We also experimented with the Transformer
model \cite{Vaswani17} but the results were worse. This could be due to a
limited grid search we performed as Transformer is known to be very sensitive
to hyperparameters.}

After a small grid search, we decided to inherit most of the hyperparameters of
the model from the best results achieved in \cite{Wang18} where
\LaTeX{}-to-Mizar translation is learned. We used relatively small LSTM cells
consisting of 2 layers with 128 units. The ``scaled Luong'' version of the
attention mechanism was used, as well as dropout with rate equal $0.2$. The
number of training steps was $10000$. (This setting was used for all our
experiments described below.)

\subsection{AIM data set}

First, NMT models were trained for each of the 20 rewrite rules in the AIM data
set. It turned out that the models, as long as the number of examples was
greater than $1000$, were able to learn the rewriting task very well, reaching
$90\%$ of accuracy on separated test sets.  This means that the task of
applying a single rewrite step seems relatively easy to learn by NMT. See Table
\ref{tab:aim_results} for all the results.

We also run an experiment on the joint set of all rewrite rules (consisting of
41396 examples). Here the task was more difficult as a model needed not only to
apply rewriting correctly, but also choose ``the right'' rewrite rule
applicable for a given term. Nevertheless, the performance was also very good,
reaching $83\%$ accuracy.

\begin{table}[]
	\caption{Results of experiments with AIM data.
	(Names of the rules correspond to folder names in the Github repo.)}
\label{tab:aim_results}
	\begin{tabular}{l|l|l|l}
		Rule: &
		\thead{Training \\ examples:} &
		\thead{Test \\ examples:} &
		\thead{Accuracy \\ on test:} \\
		\hline
		\texttt{abstrused1u}&2472&1096&86.5\%\\
		\texttt{abstrused2u}&2056&960&89.2\%\\
		\texttt{abstrused3u}&1409&666&84.3\%\\
		\texttt{abstrused4u}&1633&743&87.4\%\\
		\texttt{abstrused5u}&2561&1190&89.5\%\\
		\texttt{abstrused6u}&81&40&12.5\%\\
		\texttt{abstrused7u}&76&37&0.0\%\\
		\texttt{abstrused8u}&79&39&2.5\%\\
		\texttt{abstrused9u}&1724&817&86.7\%\\
		\texttt{abstrused10u}&3353&1573&82.9\%\\
		\texttt{abstrused11u}&10230&4604&79.0\%\\
		\texttt{abstrused12u}&7201&3153&87.2\%\\
		\texttt{instused1u}&198&97&20.6\%\\
		\texttt{instused2u}&196&87&25.2\%\\
		\texttt{instused3u}&83&41&29.2\%\\
		\texttt{instused4u}&105&47&2.1\%\\
		\texttt{instused5u}&444&188&59.5\%\\
		\texttt{instused6u}&1160&531&87.5\%\\
		\texttt{instused7u}&307&144&13.8\%\\
		\texttt{instused8u}&116&54&3.7\%\\
		\textit{union of all}&41396&11826&83.2\%
	\end{tabular}
\end{table}

\subsection{Polynomial data set}

Then experiments on more challenging but also much larger data sets for polynomial
normalization were performed. Depending on the difficulty of the data, accuracy
on the test sets achieved in our experiments varied between $70\%$ and $99\%$.
The results in terms of accuracy are shown in Table \ref{tab:poly_results}.

This high performance of the model encouraged a closer inspection of the
results.  First, we checked if in the test sets there are input examples which
differ from these in training sets only by renaming of variables. Indeed, for
each of the data sets in test sets are $5 - 15 \%$ of such ``renamed''
examples. After filtering them out, the measured accuracy drops -- but only by
$1 - 2 \%$.

An examination of the examples wrongly rewritten by the model was done. It
turns out that the wrong outputs almost always parse (in $97 - 99 \%$ of cases,
they are legal polynomial terms). Notably, depending on the difficulty of the
data set, as much as $18 - 64 \%$ of incorrect outputs are wrong only with
respect to the constants in the terms. (Typically, the NMT model proposes too
low constants compared to the correct ones.) Below $1 \%$ of wrong outputs is
correct modulo variable renaming.

\begin{table}[]
	\caption{Chosen results of experiments with polynomials. \\
	(Characteristic of formulas concerns the \textit{input} polynomials.
	Labels of the data sets correspond to folder names in the Github repo.)}
\label{tab:poly_results}
\begin{tabular}{l|l|l|c|c}
	Label &
	\thead{Function \\ symbols} &
	\thead{Constant \\ symbols} &
	\thead{Num. of \\ variables} &
	%\thead{Max \\ length} &
	\thead{Accuracy \\ on test} \\
	\hline
	\texttt{poly1} & $+$, $*$ & $0, 1$ & $1$ & $99.2\%$ \\
	\texttt{poly2} & $+$, $*$ & $0, 1$ & $2$ & $97.4\%$ \\
	\texttt{poly3} & $+$, $*$ & $0, 1$ & $3$ & $88.2\%$ \\
	\texttt{poly4} & $+$, $*$ & $0, 1, 2, 3, 4, 5$ & $5$ & $83.4\%$ \\
	\texttt{poly5} & $+$, $*$, \^{} & $0, 1$ & $2$ & $85.5\%$ \\
	\texttt{poly6} & $+$, $*$, \^{} & $0, 1, 2$ & $3$ & $71.8\%$
\end{tabular}
\end{table}

\section{Conclusions and future work}
\label{sect:conc}

NMT is not typically applied to symbolic problems, but surprisingly, it
performed very well for both described tasks. The first one was easier in terms
of complexity of the rewriting (only one application of a rewrite rule was
performed), but the number of examples was quite limited. The second task
involved more difficult rewriting -- multiple different rewrite steps were
performed to construct the examples. Nevertheless, provided many examples, NMT
could learn normalizing polynomials.

We hope this work provides a baseline and inspiration for continuing this line of
research. We see several interesting directions this work can be extended.

Firstly, more interesting and difficult rewriting problems need to be provided
for better delineation of the strength of the neural models. The described data
are relatively simple and with no direct relevance to the real unsolved
symbolic problems. But the results on these simple problems are encouraging
enough to try with more challenging ones, related to real difficulties -- e.g.,
these from TPDB data
base.\footnote{\url{http://termination-portal.org/wiki/TPDB}}

Secondly, we are going to develop and test new kinds of neural models tailored
for the problem of comprehending symbolic expressions. Specifically, we are
going to implement an approach based on the idea of TreeNN, which may be
another effective approach for this kind of tasks \cite{Evans18, Miao18,
Chakraborty18}. TreeNNs are built recursively from \textit{modules}, where the
modules correspond to parts of symbolic expression (symbols) and the shape of
the network reflects the parse tree of the processed expression. This way model
is explicitly informed on the exact structure of the expression, which in case
of formal logic is always unambiguous and easy to extract. Perhaps this way the
model could learn more efficiently from examples (and achieve higher results
even on the small AIM data sets). The authors have a positive experience of
applying TreeNNs to learn remainders of arithmetical expressions modulo small
natural numbers -- TreeNNs outperformed here neural models based on LSTM cells,
giving almost perfect accuracy. However, this is unclear how to translate this
TreeNN methodology to the tasks with the structured output, like the symbolic
rewriting task.

Thirdly, there is an idea of integrating neural rewriting architectures into
the larger systems for automated reasoning. This can be motivated by the
interesting contrast between some simpler ILP systems suffering from the
combinatorial explosion in the presence of a large number of examples and
neural methods which definitely benefit from large data sets.

We hope that this work will inspire and trigger a discussion on the above
(and other) ideas.

\section{Update: Integration data sets from Facebook Research experiments}

The work of \citet{Lample20} joined recently the line of research pursued by us here:\footnote{The first version of our work was submitted to AITP'19 in December 2018~\cite{PCUK19} and presented several times at workshops and conferences in the first half of 2019. See e.g. invited talks at SAT'19 \url{http://grid01.ciirc.cvut.cz/~mptp/sat19.pdf} and FORMAL'19 \url{http://grid01.ciirc.cvut.cz/~mptp/formal19.pdf}.} applying
sequence-to-sequence neural architectures to symbolic rewriting tasks. In case of
\cite{Lample20} the symbolic tasks are integration of a chosen set of functions (basically polynomials plus trigonometry, exp and log) and solving a class of differential equations. These tasks are less abstract than rewriting in the AIM loops theory
and involve some more operations than our polynomial dataset. The datasets are orders of magnitude larger than ours.
% are
% more difficult and more useful in
% practice:

The training examples were generated by a randomized procedure, similarly to
our polynomial dataset. The neural model used there -- Transformer
\cite{Vaswani17} -- is also a non-modified architecture originally designed for
the neural machine translation. A step in their pipe-line was
preprocessing the symbolic expressions by translating them to more compact
prefix notation. This preprocessing step was used by us in the first version
of our earlier work\footnote{Submitted to IJCAI'19.} on guiding theorem provers
by recurrent neural networks~\cite{DBLP:journals/corr/abs-1905-07961}. No
experimental justification of usefulness of prefix notation is provided in
\cite{Lample20}. We have found in another work that this is not always
beneficial \cite{Piotrowski20}.)

The performance of the trained models in \cite{Lample20} was quite high,
reaching 99\%.  Here we analyze closer this experiment and compare with our methods and datasets. First, we
wanted to see how the relatively small models used originally by us perform on
these data.  We took 3 data sets related to various kinds of integration
operations (BWD, FWD, IBP) used in \cite{Lample20} and applied the NMT model
with exactly the same hyperparameters as described in \ref{sect:exp}.
Additionally, we trained an NMT model implemented in OpenNMT, leaving all the
hyperparameters in their default settings.\footnote{In particular: the number
of training steps was 100000, the number of layers in the encoder and the
decoder was 2, the number of units in the encoder and decoder was 500, the
``scaled Luong'' version of the attention mechanism was used; the predictions
were generated using beam search of width 10 and 1 best output was considered
only.}  Table \ref{tab:fb_data_accuracy} shows the accuracy of the trained
models on the test sets, along with the reported accuracies of the Transformer
model.

\begin{table}
\centering
	\caption{Prediction accuracies of the three NMT models -- the model used in
	the previous experiments in this work, the default OpenNMT model, and the
	Transformer model used in \cite{Lample20} -- on the three integration data
	sets.}
\label{tab:fb_data_accuracy}
\begin{tabular}{rccc}
	Data set &
	\thead{small \\ NMT} &
	\thead{default \\ OpenNMT} &
	Transformer \\
	\hline
	\texttt{BWD} & 18.4\% & 67.7\% & 99.6\% \\
	\texttt{FWD} & 14.8\% & 55.7\% & 97.2\% \\
	\texttt{IBP} & 17.7\% & 64.5\% & 99.3\% \\
\end{tabular}
\end{table}

We see that the small NMT model performs much worse than the Transformer model
from  \cite{Lample20}, while the performance of the default OpenNMT model is
already quite good. However, the Transformer model in \cite{Lample20} was much
larger.  The authors do not report on how long the model was trained and what
infrastructure was used. Our small NMT model was trained for about one hour on
one GPU and our default OpenNMT model was trained for two hours on two GPUs.
With such straightforward, unmodified training procedure and short training
times the achieved performance may still be seen as surprisingly high.  By
tuning the hyperparameters and increasing the number of training steps we could
likely easily increase the performance.

To get more understanding of the data, we have done a simple analysis of the
similarity between the training and testing sets. We substituted all the
constants (i.e., digits) with \texttt{CONST} token and checked how many such
modified testing examples appear in the training examples, and how many are unique to the testing set.
We did this for all the polynomial data sets and the integration data
sets. For the latter, we also substituted plus sign for all the minus signs to
ignore the sign of integers when comparing the examples. (In the polynomial data
there is no negative integers.) The results of such analysis are shown in Table
\ref{tab:unique}.

\begin{table}
\centering
	\caption{Number of unique (not overlapping with the training set) testing examples \textit{modulo constant} or
	\textit{modulo constant and sign}. All the polynomial data sets and the
	integration data sets from \cite{Lample20} were checked.}
\label{tab:unique}
\begin{tabular}{rccc}
	Data set &
	\thead{ \# unique \\ mod. constant} &
	\thead{\# unique \\ mod. constant \\ and sign} &
	\thead{\# all test \\ examples } \\
  \hline
	\texttt{BWD}   & 7421 (80\%) & 6999 (75\%) & 9319 \\
	\texttt{FWD}   & 4404 (44\%) & 3497 (35\%) & 9986 \\
	\texttt{IBP}   & 2345 (30\%) & 1895 (24\%) & 7777 \\
	\texttt{poly1} & 34877 (58\%) & -- & 60000  \\
	\texttt{poly2} & 69160 (77\%) & -- & 90000  \\
	\texttt{poly3} & 82680 (92\%) & -- & 90000  \\
	\texttt{poly4} & 77225 (86\%) & -- & 90000  \\
	\texttt{poly5} & 79185 (88\%) & -- & 90000  \\
	\texttt{poly6} & 77764 (86\%) & -- & 90000  \\
\end{tabular}
\end{table}

We see that for some of the integration data sets the number of examples unique
modulo constant (or constant and sign) may be as low as 24\% (\texttt{IBP}).
This means that for this data set 75\% of the testing examples appear in the
training set as very similar expressions, just with changed constants and their
signs.
This motivates the need for more careful/complex
analysis of the performance of machine learning models in new domains like the
symbolic rewriting. Reporting only accuracy on the separated testing set may
not be enough. It may happen that the performance of the model is dependent on
some hidden factors, e.g., undesired ``leaks'' in the data.

The analysis done here is
initial and more detailed examination would be
useful to measure the level of generalization and memorization done by the neural models used.
Our experiments on the polynomial data show that just increasing the number of constants and variables from two to three
between poly5 and poly6 decreased the testing performance from 85.5\% to 71.8\%.
Similar ablations should be done with related experiments.
% One could for example test the
Methods such as decreasing the size of the training set, making it on average more diverse, and
measuring the average Levenshtein distance between
the training and testing examples~\cite{Wang18} are relatively straightforward to apply.
% However, our goal here is mainly just to
% motivate some discussion on the topic.

\section*{Acknowledgements}
Piotrowski was supported
by the grant of National Science Center, Poland, no. 2018/29/N/ST6/02903,
and by the European Agency COST action CA15123.
Urban and Brown were supported
by the ERC Consolidator grant no.~649043 \textit{AI4REASON}
and by the Czech project \textit{AI\&Reasoning} CZ.02.1.01/0.0/0.0/15\_003/0000466
and the European Regional Development Fund.
Kaliszyk was supported by ERC Starting grant no.~714034 \textit{SMART}.

\bibliography{ate11,references-bartosz}

\begin{thebibliography}{18}
\providecommand{\natexlab}[1]{#1}
\providecommand{\url}[1]{\texttt{#1}}
\expandafter\ifx\csname urlstyle\endcsname\relax
  \providecommand{\doi}[1]{doi: #1}\else
  \providecommand{\doi}{doi: \begingroup \urlstyle{rm}\Url}\fi

\bibitem[Bancerek et~al.(2018)Bancerek, Naumowicz, and Urban]{BancerekNU18}
Bancerek, G., Naumowicz, A., and Urban, J.
\newblock System description: {XSL}-based translator of {Mizar to LaTeX}.
\newblock In Rabe, F., Farmer, W.~M., Passmore, G.~O., and Youssef, A. (eds.),
  \emph{Intelligent Computer Mathematics - 11th International Conference,
  {CICM} 2018, Hagenberg, Austria, August 13-17, 2018, Proceedings}, volume
  11006 of \emph{Lecture Notes in Computer Science}, pp.\  1--6. Springer,
  2018.
\newblock ISBN 978-3-319-96811-7.
\newblock \doi{10.1007/978-3-319-96812-4\_1}.
\newblock URL \url{https://doi.org/10.1007/978-3-319-96812-4\_1}.

\bibitem[Chakraborty et~al.(2018)Chakraborty, Allamanis, and
  Ray]{Chakraborty18}
Chakraborty, S., Allamanis, M., and Ray, B.
\newblock Tree2tree neural translation model for learning source code changes.
\newblock \emph{CoRR}, abs/1810.00314, 2018.
\newblock URL \url{http://arxiv.org/abs/1810.00314}.

\bibitem[Evans et~al.(2018)Evans, Saxton, Amos, Kohli, and
  Grefenstette]{Evans18}
Evans, R., Saxton, D., Amos, D., Kohli, P., and Grefenstette, E.
\newblock Can neural networks understand logical entailment?
\newblock In \emph{International Conference on Learning Representations}, 2018.
\newblock URL \url{https://openreview.net/forum?id=SkZxCk-0Z}.

\bibitem[Kaliszyk et~al.(2014)Kaliszyk, Urban, Vysko\v{c}il, and
  Geuvers]{KaliszykUVG14}
Kaliszyk, C., Urban, J., Vysko\v{c}il, J., and Geuvers, H.
\newblock Developing corpus-based translation methods between informal and
  formal mathematics: Project description.
\newblock In Watt, S.~M., Davenport, J.~H., Sexton, A.~P., Sojka, P., and
  Urban, J. (eds.), \emph{Intelligent Computer Mathematics - International
  Conference, {CICM} 2014, Coimbra, Portugal, July 7-11, 2014. Proceedings},
  volume 8543 of \emph{LNCS}, pp.\  435--439. Springer, 2014.
\newblock ISBN 978-3-319-08433-6.
\newblock \doi{10.1007/978-3-319-08434-3_34}.
\newblock URL \url{http://dx.doi.org/10.1007/978-3-319-08434-3_34}.

\bibitem[Kaliszyk et~al.(2015)Kaliszyk, Urban, and Vysko\v{c}il]{KaliszykUV15}
Kaliszyk, C., Urban, J., and Vysko\v{c}il, J.
\newblock Learning to parse on aligned corpora (rough diamond).
\newblock In Urban, C. and Zhang, X. (eds.), \emph{Interactive Theorem Proving
  - 6th International Conference, {ITP} 2015, Nanjing, China, August 24-27,
  2015, Proceedings}, volume 9236 of \emph{Lecture Notes in Computer Science},
  pp.\  227--233. Springer, 2015.
\newblock ISBN 978-3-319-22101-4.
\newblock \doi{10.1007/978-3-319-22102-1_15}.
\newblock URL \url{http://dx.doi.org/10.1007/978-3-319-22102-1_15}.

\bibitem[Kaliszyk et~al.(2017)Kaliszyk, Urban, and Vyskocil]{KaliszykUV17}
Kaliszyk, C., Urban, J., and Vyskocil, J.
\newblock Automating formalization by statistical and semantic parsing of
  mathematics.
\newblock In Ayala{-}Rinc{\'{o}}n, M. and Mu{\~{n}}oz, C.~A. (eds.),
  \emph{Interactive Theorem Proving - 8th International Conference, {ITP} 2017,
  Bras{\'{\i}}lia, Brazil, September 26-29, 2017, Proceedings}, volume 10499 of
  \emph{Lecture Notes in Computer Science}, pp.\  12--27. Springer, 2017.
\newblock ISBN 978-3-319-66106-3.
\newblock \doi{10.1007/978-3-319-66107-0\_2}.
\newblock URL \url{https://doi.org/10.1007/978-3-319-66107-0\_2}.

\bibitem[Kinyon et~al.(2013)Kinyon, Veroff, and Vojtechovsk{\'{y}}]{Veroff13}
Kinyon, M.~K., Veroff, R., and Vojtechovsk{\'{y}}, P.
\newblock Loops with abelian inner mapping groups: An application of automated
  deduction.
\newblock In Bonacina, M.~P. and Stickel, M.~E. (eds.), \emph{Automated
  Reasoning and Mathematics - Essays in Memory of William W. McCune}, volume
  7788 of \emph{Lecture Notes in Computer Science}, pp.\  151--164. Springer,
  2013.
\newblock ISBN 978-3-642-36674-1.
\newblock \doi{10.1007/978-3-642-36675-8\_8}.
\newblock URL \url{https://doi.org/10.1007/978-3-642-36675-8\_8}.

\bibitem[Lample \& Charton(2020)Lample and Charton]{Lample20}
Lample, G. and Charton, F.
\newblock Deep learning for symbolic mathematics.
\newblock In \emph{8th International Conference on Learning Representations,
  {ICLR} 2020, Addis Ababa, Ethiopia, April 26-30, 2020}. OpenReview.net, 2020.
\newblock URL \url{https://openreview.net/forum?id=S1eZYeHFDS}.

\bibitem[Luong et~al.(2017)Luong, Brevdo, and Zhao]{Luong17}
Luong, M., Brevdo, E., and Zhao, R.
\newblock Neural machine translation (seq2seq) tutorial.
\newblock \emph{https://github.com/tensorflow/nmt}, 2017.

\bibitem[Miao et~al.(2018)Miao, Wang, Le, Tao, Shang, Yan, and Zhao]{Miao18}
Miao, N., Wang, H., Le, R., Tao, C., Shang, M., Yan, R., and Zhao, D.
\newblock Tree2tree learning with memory unit, 2018.
\newblock URL \url{https://openreview.net/forum?id=Syt0r4bRZ}.

\bibitem[Muggleton \& De~Raedt(1994)Muggleton and
  De~Raedt]{muggleton1994inductive}
Muggleton, S. and De~Raedt, L.
\newblock Inductive logic programming: Theory and methods.
\newblock \emph{The Journal of Logic Programming}, 19:\penalty0 629--679, 1994.

\bibitem[Piotrowski \& Urban(2019)Piotrowski and
  Urban]{DBLP:journals/corr/abs-1905-07961}
Piotrowski, B. and Urban, J.
\newblock Guiding theorem proving by recurrent neural networks.
\newblock \emph{CoRR}, abs/1905.07961, 2019.
\newblock URL \url{http://arxiv.org/abs/1905.07961}.

\bibitem[Piotrowski \& Urban(2020)Piotrowski and Urban]{Piotrowski20}
Piotrowski, B. and Urban, J.
\newblock Stateful premise selection by recurrent neural networks.
\newblock \emph{CoRR}, abs/2004.08212, 2020.
\newblock URL \url{https://arxiv.org/abs/2004.08212}.

\bibitem[Piotrowski et~al.(2019)Piotrowski, Brown, Urban, and Kaliszyk]{PCUK19}
Piotrowski, B., Brown, C., Urban, J., and Kaliszyk, C.
\newblock Can neural networks learn symbolic rewriting?
\newblock In \emph{Proceedings of AITP 2019}, 2019.
\newblock URL \url{http://aitp-conference.org/2019/abstract/paper\%2014.pdf}.

\bibitem[Sutcliffe(2017)]{Sutcliffe17}
Sutcliffe, G.
\newblock {The TPTP Problem Library and Associated Infrastructure. From CNF to
  TH0, TPTP v6.4.0}.
\newblock \emph{Journal of Automated Reasoning}, 59\penalty0 (4):\penalty0
  483--502, 2017.

\bibitem[Urban(1998)]{urban98}
Urban, J.
\newblock Experimenting with machine learning in automatic theorem proving.
\newblock Master's thesis, Faculty of Mathematics and Physics, Charles
  University, Prague, 1998.
\newblock English summary at
  \url{https://www.ciirc.cvut.cz/~urbanjo3/MScThesisPaper.pdf}.

\bibitem[Vaswani et~al.(2017)Vaswani, Shazeer, Parmar, Uszkoreit, Jones, Gomez,
  Kaiser, and Polosukhin]{Vaswani17}
Vaswani, A., Shazeer, N., Parmar, N., Uszkoreit, J., Jones, L., Gomez, A.~N.,
  Kaiser, L., and Polosukhin, I.
\newblock Attention is all you need.
\newblock In Guyon, I., von Luxburg, U., Bengio, S., Wallach, H.~M., Fergus,
  R., Vishwanathan, S. V.~N., and Garnett, R. (eds.), \emph{Advances in Neural
  Information Processing Systems 30: Annual Conference on Neural Information
  Processing Systems 2017, 4-9 December 2017, Long Beach, CA, {USA}}, pp.\
  6000--6010, 2017.
\newblock URL \url{http://papers.nips.cc/paper/7181-attention-is-all-you-need}.

\bibitem[Wang et~al.(2018)Wang, Kaliszyk, and Urban]{Wang18}
Wang, Q., Kaliszyk, C., and Urban, J.
\newblock First experiments with neural translation of informal to formal
  mathematics.
\newblock In Rabe, F., Farmer, W.~M., Passmore, G.~O., and Youssef, A. (eds.),
  \emph{11th International Conference on Intelligent Computer Mathematics (CICM
  2018)}, volume 11006 of \emph{LNCS}, pp.\  255--270. Springer, 2018.
\newblock \doi{10.1007/978-3-319-96812-4\_22}.
\newblock URL \url{https://doi.org/10.1007/978-3-319-96812-4\_22}.

\end{thebibliography}
\bibliographystyle{icml2019}

\end{document}